\documentclass[runningheads]{llncs}
\usepackage[T1]{fontenc}
\usepackage{graphicx}
\usepackage{amsmath,amssymb}
\usepackage{dsfont}
\usepackage{array}
\usepackage{booktabs}
\usepackage{multirow}
\usepackage{subcaption}
\usepackage{caption}
\usepackage{bm}
\usepackage{mathrsfs}
\usepackage[colorlinks=true, linkcolor=blue, citecolor=blue, urlcolor=blue]{hyperref}
\usepackage{color}

\usepackage{orcidlink}
\usepackage{color}

\newcommand{\argmax}{\mathop{\mathrm{argmax}}\limits}
\newcommand{\argmin}{\mathop{\mathrm{argmin}}\limits}

\DeclareMathOperator{\argmaxinline}{argmax}

\usepackage[backend=biber,style=numeric]{biblatex}
\makeatletter
\newcommand\blfootnote[1]{%
  \begingroup
  \renewcommand\thefootnote{}
  \footnote{#1}
  \addtocounter{footnote}{-1}
  \endgroup
}
\makeatother

\begin{document}
\title{Uncertainty-Aware Concept Bottleneck Models with Enhanced Interpretability}
\titlerunning{Uncertainty-Aware CBMs with Enhanced Interpretability}
%
\author{Haifei Zhang\inst{1}\orcidlink{0000-0003-4488-1631} \and
Patrick Barry\inst{1}\orcidlink{0009-0003-0715-5478} \and
Eduardo Brandao\inst{1,2}\orcidlink{0000-0002-7146-8255}
}

\authorrunning{H. Zhang et al.}


\institute{%
  Université Jean Monnet Saint-Etienne, Institut d’Optique Graduate School, Laboratoire Hubert Curien UMR-CNRS 5516, Saint-Etienne, France \\
  \email{\{haifei.zhang, eduardo.brandao\}@univ-st-etienne.fr}\\
  \email{\{patrick.barry\}@etu.univ-st-etienne.fr}
\and
CNRS, Inria}

\maketitle

\blfootnote{This paper has been accepted for the Workshop \href{https://project.inria.fr/aimlai/program/}{AIMLAI} at ECML-PKDD 2025.}

\begin{abstract}
In the context of image classification, Concept Bottleneck Models (CBMs) first embed images into a set of human-understandable concepts, followed by an intrinsically interpretable classifier that predicts labels based on these intermediate representations. While CBMs offer a semantically meaningful and interpretable classification pipeline, they often sacrifice predictive performance compared to end-to-end convolutional neural networks. Moreover, the propagation of uncertainty from concept predictions to final label decisions remains underexplored. In this paper, we propose a novel uncertainty-aware and interpretable classifier for the second stage of CBMs. Our method learns a set of binary class-level concept prototypes and uses the distances between predicted concept vectors and each class prototype as both a classification score and a measure of uncertainty. These prototypes also serve as interpretable classification rules, indicating which concepts should be present in an image to justify a specific class prediction. The proposed framework enhances both interpretability and robustness by enabling conformal prediction for uncertain or outlier inputs based on their deviation from the learned binary class-level concept prototypes.

\keywords{Explainable AI  \and Concept bottleneck models \and Prototypes \and Uncertainty \and Robustness.}
\end{abstract}

\section{Introduction}

Deep neural networks have achieved remarkable success in computer vision tasks, yet their lack of interpretability remains a significant concern \cite{lipton2018mythos}. This issue becomes especially critical in high-stakes domains such as medical diagnosis and fraud detection, where understanding model decisions is essential \cite{rudin2022interpretable}. A large number of post-hoc explanation methods have been proposed to provide insights into already-trained deep neural networks, such as various gradient-based saliency maps \cite{antamis2024xdnn,hosain2024xdl,sundararajan2017IG}. However, such pixel-level heatmaps often fail to offer high-level semantic understanding, limiting their interpretability \cite{kim2018TCAV}.

Recently, Concept Bottleneck Models (CBMs) have emerged as a promising approach for improving interpretability in image classification. In CBMs, an image is first embedded into a set of human-understandable intermediate concepts, and then a downstream interpretable classifier is trained to predict the final label based on these concepts \cite{koh2020CBM}. The concept embedding stage often uses a convolutional neural network (CNN) to produce a confidence score $\hat{c}_k \in  [0,1]$ for the presence of each concept. For the second stage, which maps concept predictions to labels, simple models such as logistic regression (LR) or shallow multi-layer perceptrons (MLPs) are commonly used \cite{koh2020CBM}.

This second stage can be viewed as a tabular-data classification problem: predicting labels from concept activation vectors. While using simple models enables clear attribution of concept importance and supports interpretability, it also comes at the cost of limited expressive power \cite{vincenzi2016tradeoff}. Logistic regression, in particular, may fail to capture complex concept interactions and is well known to perform sub-optimally compared to more flexible classifiers. Thus, designing interpretable yet more accurate alternatives remains an important and underexplored direction.

Another critical issue in the CBM pipeline is the propagation of uncertainty from the concept embedding stage to the final label prediction. This becomes especially problematic when the predicted concepts are noisy or ambiguous \cite{kim2023probabilistic,vandenhirtz2024stochastic}. In practice, some images may result in very sparse concept activation vectors, while others may produce unusually dense patterns. These atypical activations can be viewed as outliers, especially for unseen images. Due to the limitations of CNNs, the prediction of concepts may also be wrong for ambiguous images. However, deterministic classifiers like logistic regression will still produce a label prediction, even in the presence of severe concept uncertainty, which would be dangerous in high-stakes domains. Despite the practical importance of modeling uncertainty in this stage, it remains largely unaddressed in current literature.

\begin{figure}[t]
    \centering
    \includegraphics[width=\linewidth]{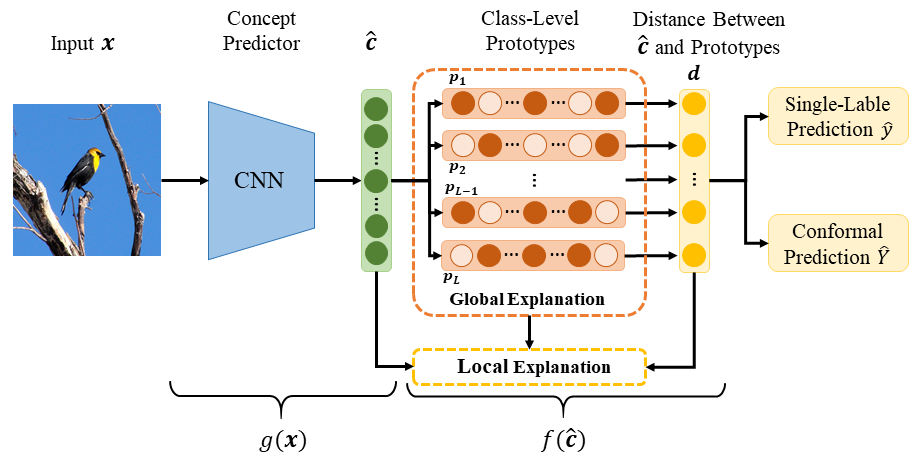}
    \caption{Model structure of our proposed class-level prototype network.}
    \label{fig:model-structure}
\end{figure}

Motivated by these two challenges, we propose a novel class-level binary prototype network as an uncertainty-aware and interpretable classifier for the second stage of CBMs, as illustrated in Fig.~\ref{fig:model-structure}. Specifically, we learn one binary-valued prototype per class, representing the ideal set of concept activations associated with that class. We then compute distances between predicted concept vectors and class prototypes as both a prediction metric and a measure of uncertainty. Based on this uncertainty, our model is convenient for conducting conformal predictions for set-valued predictions, as well as outlier detection. Furthermore, for interpretation, these learned prototypes are naturally classification rules for each class as the global explanation of the model. The concept-wise difference between concept prediction and class prototypes can serve as natural local explanations for particular instances and predictions.

The rest of the paper is structured as follows. Section~\ref{sec:2} provides preliminaries about CBMs. We present our proposed model in Section~\ref{sec:3} and evaluate it in Section~\ref{sec:4}. Finally, a conclusion is drawn in Section~\ref{sec:5}.

\section{Concept bottleneck models}\label{sec:2}

Let $\boldsymbol{x} \in \mathcal{X}$ denote the raw input image, $\boldsymbol{c} \in \mathcal{C} = \{c_1, \dots, c_K\} \subseteq \{0,1\}^K$ represent the associated concept vector consisting of $K$ human-understandable concepts, and $y \in \mathcal{Y}=\{y_1, \dots, y_L\}$ be the target label. A CBM consists of learning two components \cite{koh2020CBM}:

\begin{itemize}
    \item A concept predictor $g$: $\mathcal{X} \to [0,1]^K$ that maps the input image $\boldsymbol{x}$ to a soft concept vector $\boldsymbol{\hat{c}} = g(x)$, where each element represents the confidence that the corresponding concept is observed in the input image \footnote{The prediction of concepts can also be hard binary values or even raw logits. In this paper, we focus on soft concept confidence scores.}.

    \item A label predictor $f$: $[0,1]^K \to \mathcal{Y}$, which maps the predicted concept vector $\boldsymbol{\hat{c}}$ to the final label $\hat{y} = f(\boldsymbol{\hat{c}})$.
\end{itemize}

During the training phase, a dataset of $N$ samples $\{(\boldsymbol{x}^{(i)},\ \boldsymbol{c}^{(i)},\ y^{(i)})\}_{i=1}^N$ is required. In most cases, the concept annotation $\boldsymbol{c}$ is binary (indicating the presence or absence of each concept). There are three typical training strategies to learn both $g$ and $f$ \cite{koh2020CBM}. 

\begin{enumerate}
    \item \textbf{Independent training}: Here, the concept predictor $g$ is trained on input images and concept annotations. The label predictor $f$ is trained separately on the concept annotations and the ground-truth labels,
    \begin{subequations}
        \begin{align}
            g^{*} = \argmin_{g} \frac{1}{N} \sum_{i=1}^N \mathcal{L}_C (g(\boldsymbol{x}^{(i)}),\ \boldsymbol{c}^{(i)}), \label{eq:1a} \\
            \ f^{*} = \argmin_{f} \frac{1}{N} \sum_{i=1}^N \mathcal{L}_Y(f(\boldsymbol{c}^{(i)}),\ y^{(i)}), \label{eq:1b}
        \end{align}
    \end{subequations}
    where $\mathcal{L}_C$ and $\mathcal{L}_Y$ denote the loss functions for concept prediction and label prediction, respectively. While concept annotations are used to train $f$, we still use the predicted concepts (scores) from $g^*$ to make label predictions in the inference phase. i.e., $\hat{y} = f^{*}(g^{*}(\boldsymbol{x}))$.

    \item \textbf{Sequential training}: The concept predictor $g$ is trained first using concept supervision only. Once trained, $g^*$ is frozen, and the label predictor $f$ is trained sequentially on the predicted concept vectors given  by $g^*$ and the ground-truth labels,
    \begin{subequations}
        \begin{align}
            g^{*} &= \argmin_{g} \frac{1}{N} \sum_{i=1}^N \mathcal{L}_C (g(\boldsymbol{x}^{(i)}),\ \boldsymbol{c}^{(i)}), \label{eq:2a} \\
            f^{*} &= \argmin_{f} \frac{1}{N} \sum_{i=1}^N \mathcal{L}_Y(f(g^{*}(\boldsymbol{x}^{(i)})),\ y^{(i)}). \label{eq:2b}
        \end{align}
    \end{subequations}

    \item \textbf{Joint training}: Both $g$ and $f$ are optimized simultaneously with respect to a combined loss function,
    \begin{equation}\label{eq:3}
        g^{*},\, f^{*} = \argmin_{g,\, f} \frac{1}{N}\sum_{i=1}^N\left[ \mathcal{L}_Y(f(g(\boldsymbol{x}^{(i)})),\ y^{(i)}) + \lambda \mathcal{L}_C (g(\boldsymbol{x}^{(i)}),\ \boldsymbol{c}^{(i)})\right],
    \end{equation}
    where $\lambda \ge 0$ controls the trade-off between concept prediction accuracy and final label classification performance. If $\ \lambda=0$, the model will degenerate into a black-box model.
\end{enumerate}

Depending on the training strategy, different forms of concept vectors may be used for training the label predictor $f$. In Eq.~\eqref{eq:1b}, binary ground-truth concepts are used, whereas in Eq.~\eqref{eq:2b} and Eq.~\eqref{eq:3}, predicted soft concept probabilities are used instead.

These three training strategies of CBMs differ in terms of prediction accuracy, interpretability, and train-test consistency. Joint training often achieves the best classification performance due to end-to-end optimization, but it may sacrifice interpretability, as the learned concepts can drift from their intended meanings, also referred to as the information leakage phenomenon \cite{havasi2022addressing,margeloiu2021concept,parisini2025leakage}. Independent training preserves the strongest alignment between concepts and human semantics by training the label predictor on ground-truth concepts, but it tends to perform worse and suffers from train-test mismatch, since the label predictor is trained on ground-truth concept annotations but tested on predicted (and possibly noisy) concepts. Sequential training provides a compromise that does not suffer from training-testing mismatches because the label predictor is trained on predicted concepts rather than ground-truth concept annotations, and it mitigates the information leakage problem because the concept predictor is trained separately, independent of the label prediction accuracy.

This paper will focus on the second stage of the sequential training strategy to explore a novel uncertainty-aware and interpretable classifier, which will be detailed in Section~\ref{sec:3}.

\section{Class-level prototype classifier}\label{sec:3}

In this section, we present our binary class-level prototype classifier (CLPC). The proposed method not only enables accurate classification by matching predicted concept vectors to learned class prototypes but also allows for intuitive explanation by attributing uncertainty to individual concepts. We detail the design of the prototype-based classifier, its training procedure, and how it naturally supports uncertainty-awareness and interpretability in this section.

\subsection{Model structure and inference}

As illustrated in Fig.~\ref{fig:model-structure}, the proposed model consists of two main components. The first stage employs a classical convolutional neural network (CNN) backbone (e.g., ResNet50 \cite{he2016resnet} or Inception-V3 \cite{szegedy2016inceptionV3}), denoted as $g$, which encodes the input image into a concept activation vector (also referred to as concept predictions). Given an input image $\boldsymbol{x}$, the concept predictions are obtained as $\boldsymbol{\hat{c}} = g(\boldsymbol{x})$. Each predicted concept can take various forms, including raw logits, continuous confidence scores in the range $[0,1]$ via a sigmoid activation, or binary values. In this work, we adopt the continuous confidence scores as the representation of predicted concepts.

The second stage introduces our proposed class-level prototype classifier, which serves as an alternative to conventional approaches such as logistic regression or shallow multilayer perceptrons. The classifier is parameterized by a set of class prototypes, where each prototype $\boldsymbol{p}_j \in \{0,1\}^K$, matching the dimensionality of the concept vector, encodes the presence or absence of specific concepts associated with class $y_j$.

To perform classification for a given image $\boldsymbol{x}$, we compute the distance between the predicted concept vector $\boldsymbol{\hat{c}}$ and each class prototype $\boldsymbol{p}_j$. The final prediction corresponds to the class whose prototype is the closest to the predicted concept vector:

\begin{equation}\label{eq:inference}
\hat{y} = \argmin_{y_j \in \mathcal{Y}} d(\boldsymbol{\hat{c}},\ \boldsymbol{p}_j).
\end{equation}
where $d(\cdot, \cdot)$ denotes a distance metric. In this work, we adopt the Manhattan distance (L1 norm), motivated by its compatibility with sparse representations and interpretability. The distance between the predicted concept vector $\boldsymbol{\hat{c}}$ and a class prototype $\boldsymbol{p}_j$ is computed as:

\begin{equation}\label{eq:L1}
d(\boldsymbol{\hat{c}},\ \boldsymbol{p}_j) = \sum_{k=1}^K \left| \hat{c}_k - p_{jk} \right|.
\end{equation}
Under this formulation, each concept-wise term $\delta_k = \left| \hat{c}_k - p_{jk} \right|$ can be interpreted as the uncertainty or mismatch introduced by concept $c_k$ with respect to class $y_j$. This property enables a fine-grained analysis of which concepts contribute to the classification decision or potential ambiguity, enhancing the transparency and interpretability of the model.

\subsection{Class-level prototype learning}

In this work, we adopt a sequential training strategy to simplify optimization and enhance interpretability. Specifically, we decouple the training process into two stages.

We first train the CNN-based concept predictor $g$ to map input images to meaningful concept vectors. This stage is supervised using concept annotations and fine-tuning pretrained ResNet50 or Inception-V3 models, to minimize concept prediction loss $\mathcal{L}_{C}$, typically using binary cross-entropy for each concept dimension.

Once the concept predictor is trained and fixed, we train the class-level prototype classifier based on the predicted concept vectors. In this stage, the CNN parameters remain frozen, and we optimize only prototypes ${\boldsymbol{p}_j},\ j=1,\dots, L$. In order to keep smoother gradients during training, the sigmoid function $\sigma$ is applied to transform real values to approximate binary values in prototypes, i.e., $p_{jk} = \sigma(w_{jk})$, where $w_{jk} \in \mathbb{R},\ j=1,\dots, L$ and $k=1, \dots, K$. 

For a given image $\boldsymbol{x}$ with ground-truth label $y_{j^*}$, we compute its concept vector $\boldsymbol{\hat{c}} = g(\boldsymbol{x})$ and train class-level prototypes by minimizing:

\begin{equation}\label{eq:Lp}
\mathcal{L}_{p} (f(\boldsymbol{\hat{c}}),\ y_{j^*}) = d(\boldsymbol{\hat{c}},\ \boldsymbol{p}_{j^*}) - \frac{1}{L-1} \sum_{j \ne j^*} d(\boldsymbol{\hat{c}},\ \boldsymbol{p}_{j}),
\end{equation}
where $f(\boldsymbol{\hat{c}})$ returns the distances between the predicted concept vector $\boldsymbol{\hat{c}}$ and each class prototype. Minimizing this loss function encourages minimizing the distance from $\boldsymbol{\hat{c}}$ to the prototype of the ground-truth class while maximizing the distance to prototypes of other classes.

To enhance the interpretability and discriminative capacity of the learned class prototypes, we introduce two regularization terms during training: a sparsity constraint and a binarization constraint. 

The sparsity constraint encourages each prototype to activate only a small subset of concept dimensions, thereby promoting more selective and interpretable representations. This helps the model focus on the most salient concept relevant to each class, mitigates the influence of noisy or redundant dimensions, and improves generalization. We implement this by applying an $L_1$ norm penalty to each class prototype:

\begin{equation}\label{eq:Ls}
    \mathcal{L}_s = \sum_{j=1}^L ||\boldsymbol{p}_j||_1. 
\end{equation}

In addition, we impose a binarization constraint to encourage each element of the prototype to be close to either 0 or 1. This discrete-like structure provides a foundation for symbolic reasoning or rule-based interpretation. We formulate this as a penalty on deviations from binary values:

\begin{equation}\label{eq:Lb}
    \mathcal{L}_b = \sum_{j=1}^L \sum_{k=1}^K (1-p_{jk})\cdot p_{jk}.
\end{equation}

Therefore, the overall total loss function is a weighted sum of the primary prototype distance loss and the two regularization terms:

\begin{equation}\label{eq:Lt}
    \mathcal{L} = \frac{1}{N}\sum_{i=1}^N \mathcal{L}_{p} (f(\boldsymbol{\hat{c}}^{(i)}), y^{(i)}) + \lambda_s \mathcal{L}_s + \lambda_b \mathcal{L}_b,
\end{equation}
where $\lambda_s$ and $\lambda_b$ denote hyperparameters that control the influence of sparsity and binarization, respectively. These values can be tuned based on validation performance.

Finally, after training, we fix prototype values to binary ones for inference:
\begin{equation}
    p_{jk} = \mathds{1}(\sigma(w_{jk}) \ge 0.5),
\end{equation}
where $\mathds{1}(\cdot)$ is the indicator function.

\subsection{Uncertainty-aware prediction}

To enable uncertainty-aware predictions, we adopt the conformal prediction framework \cite{novello2024ood-cp,shafer2008tutorial-cp}, which provides a formal mechanism to quantify predictive uncertainty with statistical guarantees. In our prototype-based model, for a given image $\boldsymbol{x}$, the nonconformity score for each class is defined as the distance between the predicted concept vector $\boldsymbol{\hat{c}}$ and the prototype $\boldsymbol{p}_j$ of each class. Intuitively, the farther the predicted concept vector of an input is from a class prototype, the less likely it is to belong to that class.

Given a held-out calibration set $\{(\boldsymbol{x}^{(i)},\ \boldsymbol{c}^{(i)},\ y^{(i)})\}_{i=1}^{N_{\text{cal}}}$, we denote the index of $y^{(i)}$ as $j^{*(i)}$ and compute the nonconformity score for each calibration sample as:
\begin{equation}
    s^{(i)} = d(\boldsymbol{\hat{c}}^{(i)},\  \boldsymbol{p}_{j^{*(i)}}),\ i=1, \dots,N_{\text{cal}},
\end{equation}
and determine the $(1 - \alpha)$-quantile $q_{1 - \alpha}$ from the distribution of nonconformity scores. For a test input $\boldsymbol{x}$ and its predicted concept vector $\boldsymbol{\hat{c}}$, we then construct the prediction set as:
\begin{equation}
\boldsymbol{\hat{Y}} = \left\{y_j \in \mathcal{Y} \,\middle|\, d(\boldsymbol{\hat{c}},\ \boldsymbol{p}_{j}) \leq q_{1 - \alpha} \right\}.
\end{equation}
This prediction set contains all labels whose prototypes are sufficiently close to the predicted concept vector. By construction, the conformal predictor satisfies the marginal coverage guarantee:
\begin{equation}
\mathbb{P}\left[ y \in \boldsymbol{\hat{Y}} \right] \geq 1 - \alpha,
\end{equation}
under the assumption that the calibration and test data are exchangeable.

Importantly, this framework allows the model to express uncertainty in a flexible and interpretable manner. When the model is confident, the prediction set $\boldsymbol{\hat{Y}}$ may contain a single label; when the input is ambiguous or lies near class boundaries, multiple labels may be included; and when the input is far from all prototypes (e.g., outliers due to noise or distributional shift), the prediction set may become empty, effectively rejecting the input. This uncertainty-aware behavior is particularly valuable in high-risk scenarios where abstention is preferable to overconfident misclassifications.

\begin{table}[t]
    \centering
    \caption{Example of distance decomposition between predicted concept scores and the closest class prototypes}
    \renewcommand{\arraystretch}{1.5}
    \begin{tabular}{p{2.5cm}*{8}{>{\centering\arraybackslash}p{1.1cm}}}
    \toprule[1.5pt]
         & $c_1$ &  $c_2$ &  $c_3$ &  $c_4$ &  $c_5$ &  $c_6$ &  $c_7$ &  $c_8$  \\
    \midrule
         Class prototype & 1 & 1 & 0 & 1 & 0 & 1 & 0 & 0 \\
        Concept scores & 0.7 &  0.9 &  0.1 &  1 &  0 &  0.8 &  0.5 &  0.2 \\
        Distance = 1.4 & 0.3 & 0.1 & 0.1 & 0 & 0 & 0.2 & 0.5 & 0.2 \\
    \bottomrule[1.5pt]
    \end{tabular}
    \label{tab:example}
\end{table}

\begin{figure}[!ht]
    \centering
    \includegraphics[width=\linewidth]{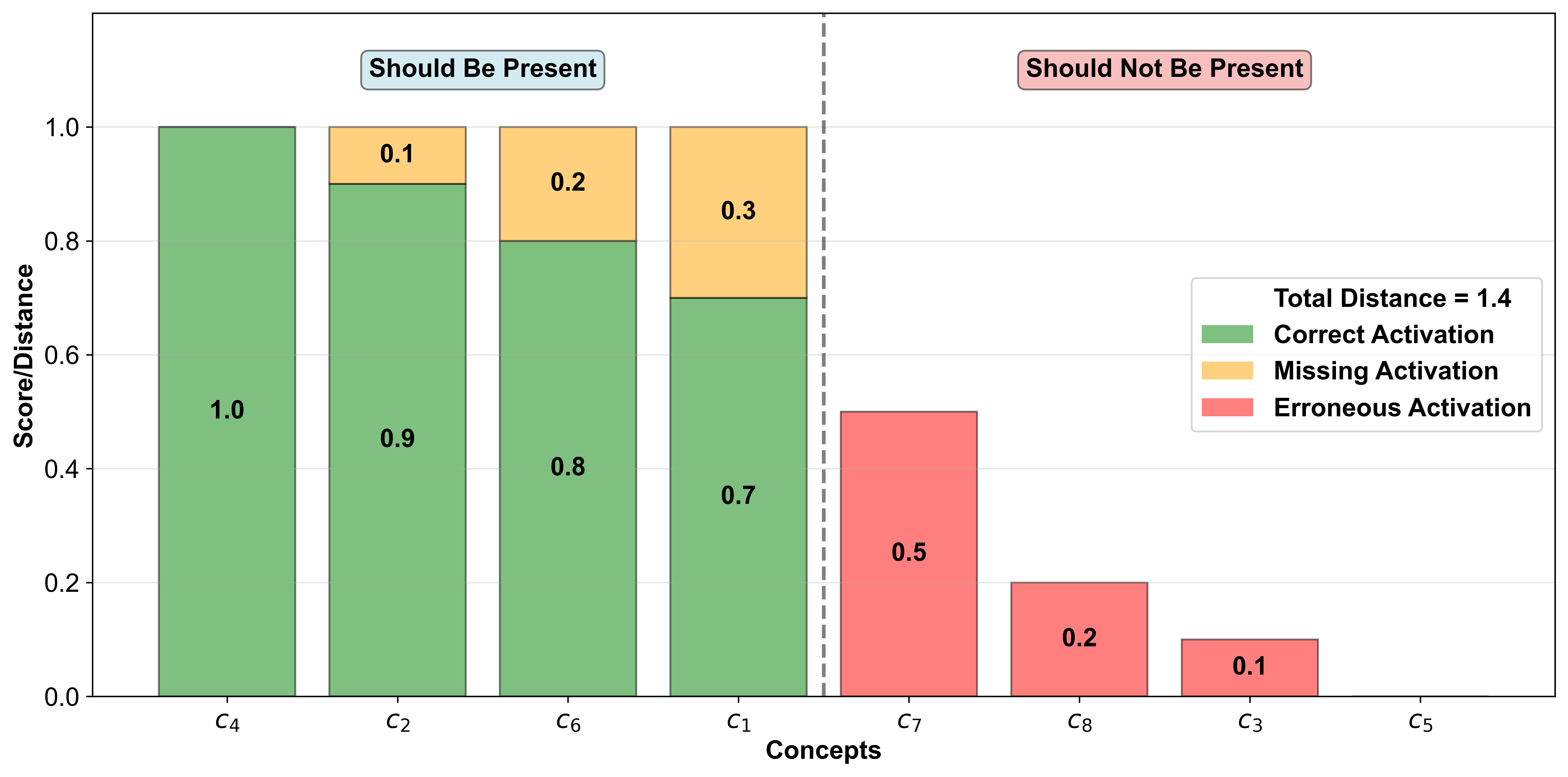}
    \caption{Visualization of distance decomposition between predicted concept scores and the closest class prototypes}
    \label{fig:decomposition-example}
\end{figure}

\subsection{Interpretability}

In our framework, each class prototype can be interpreted as a rule-like template that specifies the necessary concept-level evidence for classifying an input into that class. This framework is a combination of concept detection, classification rules, and prototype-based reasoning. Formally, a class-level prototype provides a subset of concepts that are expected to be present in a typical example of the target class. For example, such a subset of concepts for the melanoma disease could be \{\texttt{pigment network, negative network, streaks, blotches, blue white veil, regression structures}\}. This kind of learning result can be easily presented and understood by humans, requiring little basic domain knowledge. 

In contrast to logistic regression, which explains predictions through abstract feature weightings, our prototype-based model produces explanations that are simpler and more aligned with human reasoning. Humans tend to recognize categories based on the presence or absence of salient concepts, rather than by computing the exact contribution of each feature, which is supported by the prototype theory of categorization \cite{rosch1973natural}. By modeling each class as a concept prototype and classifying inputs based on their similarity to these prototypes, our approach mirrors this cognitive strategy and enhances the interpretability of model decisions.

For a specific input image and its prediction, we propose a distance decomposition visualization that attributes the distance between predicted concept scores and the closest class prototype to individual concept contributions. This approach provides insights into both why the prediction is made and which concepts contribute most to classification uncertainty.

Our visualization treats concepts differently based on their expected presence states in the prototype. For concepts that should be present (prototype value = 1), we represent each as a unit-height bar where predicted concept confidence scores are shown in green and distance contributions (gaps from ideal presence) in yellow. For concepts that should not be present (prototype value = 0), we display red bars proportional to their predicted concept scores, as any activation directly contributes to prediction uncertainty. Therefore, the total surface of the yellow and red bars is equal to the uncertainty.

Fig.~\ref{fig:decomposition-example} provides a concrete example of a class prototype, predicted concept scores, and distance decomposition based on Table~\ref{tab:example}. The visualization reveals that this input instance is classified to the class because expected concepts $c_1$, $c_2$, $c_4$, and $_6$ are detected as present. For the uncertainty, $c_1$ and $c_6$ show the most significant contributions among the required concepts. The undesired activations of $c_7$ and $c_8$ also contribute significantly to classification uncertainty. Considering all activations, the concepts on the left side of the visualization are the main factors that allow us to make predictions, while those in the middle are the ones that contribute the most to the uncertainty.

\subsection{Concept intervention}

In CBMs, misclassifications can originate from either aleatoric
uncertainty (e.g., mislabeling) or epistemic uncertainty coming from the training bias of either $f$ or $g$. However, the modular structure of CBMs allow us to intervene in concept prediction to correct misclassifications. For this problem, the order of concepts to be intervened is an important factor in the efficiency of intervention. Conventionally, concept interventions are prioritized using feature importance scores. However, such a global strategy overlooks the current prediction error in concepts. To address this, we propose a new potential gain-maximizing strategy. 

Suppose we observe a misclassified example $\boldsymbol{x}$, i.e., $\hat{y} =\argmaxinline_{y_j \in \mathcal{Y}} {f_j(\boldsymbol{\hat{c}})}$ while its true class is $y_{j^*}$. 
It should be noted that our model can be seen as maximizing the negative distance. We write $\boldsymbol{\delta} = \boldsymbol{c}^* - \boldsymbol{\hat{c}}$, the difference between the concept prediction and a hypothetical true (unknown) $\boldsymbol{c}^*$ such that $y_{j^*} =\argmaxinline_{y_j \in \mathcal{Y}} {f_j(\boldsymbol{c}^*)}$. If $f$ is differentiable, to first order in $\boldsymbol{\delta}$ we have:


\begin{equation}
f( \mathbf{c}^{ * }) = f( \hat{\mathbf{c}} + \boldsymbol{\delta}) \approx f(\hat{\mathbf{c}}) +  \sum\limits_{k=1}^K \frac{\partial f}{\partial c_k}\Big|_{\mathbf{\hat{c}}} \delta_k.
\end{equation}
One can read this expression as stating that the classifier output can be changed to match the ``true'' output by changing the predicted concept vector by $\boldsymbol{\delta}$.

In the CBM framework, we perform this concept intervention using a greedy approach. To correct a misclassification, we iteratively select and modify the concept that maximizes the following ``gain'' metric, making $\boldsymbol{\hat{c}}$ as close to the true $\boldsymbol{c}^{ * }$ as necessary, by setting $\delta_k=c^{ * }_k - \hat{c}_k$. The gain for a concept $c_k$ is defined as:
\begin{equation}
    \text{Gain}_k = \left| \frac{\partial f}{\partial c_k}\Big|_{\hat{\boldsymbol{c}}} \right| \cdot \left| c^{ * }_k - \hat{c}_k \right|.
\end{equation}
This proposal is quite general and can be applied to any differentiable model, such as logistic regression (LR), multilayer perception (MLP), and our proposed distance-based classifiers.
In particular, if $f$ is a logistic regression classifier, maximizing the posterior probability of a class is equivalent to maximizing its logits. In this case, the derivative with respect to $c_k$ is just the constant weight $w_{j^*k}$ and we can rewrite the intervention gain as $\mathrm{gain}_k = |w_{j^{*}k}|\cdot |c^{*}_k - \hat{c}_k|$.  If we impose the ansatz $c^{*}_k = \mathds{1}(w_{j^*k} > 0)$, the gain is presented as:
\begin{equation}\label{eq:lt-gain}
\text{LR-Gain}_k = w_{j^*k} \cdot \left( \mathds{1}(w_{\tilde{y}k} > 0) - \hat{c}_k \right).
\end{equation}

Otherwise, if $f$ is our proposed CLPC model, it can be rewritten as:
\begin{equation}
\hat{y} = \argmin_{y_j \in \mathcal{Y}} \sum_{k=1}^K \left| p_{jk} - \hat{c}_k\right| = \argmax_{y_j \in \mathcal{Y}} \sum_{k=1}^K -\left| p_{jk} - \hat{c}_k\right|.
\end{equation}
The value of $c^{*}_k$ is intuitively set to $p_{j^*k}$, either 0 or 1, and the absolute value of the derivative with respect to $c_k$ is always equal to 1. Therefore, the intervention gain on concept $c_k$ is 
\begin{equation}\label{eq:clpc-gain}
\text{CLPC-Gain}_k = |p_{j^{*}k} - \hat{c}_k|.
\end{equation}

Once the intervention gains on each concept are calculated, concepts are sorted in descending order of gains, and are iteratively corrected by setting $\hat{c}_k = c^*_k$. After each edit, the label prediction is recomputed, and the process stops once the prediction aligns with the target class.

This concept intervention process produces the minimal set of concept changes needed to flip the prediction and can be interpreted as a counterfactual explanation: the model would have predicted the target class if certain concepts had been corrected. This targeted intervention strategy highlights the model’s sensitivity to specific concepts, providing a powerful tool for both human-in-the-loop decision-making and post-hoc model debugging.

\section{Experiments}\label{sec:4}

We conducted experiments on three benchmark datasets to evaluate the effectiveness and robustness of our proposed Class-Level Prototype Classifier (CLPC):

\begin{itemize}
    \item \textbf{CUB-200-2011}: A fine-grained bird classification dataset containing 11788 images, annotated with 312 concepts and 200 classes.
    \item \textbf{Derm7pt}: A skin lesion dataset with 2000 images, 7 clinically meaningful concepts, and 5 diagnostic categories.
    \item \textbf{RIVAL10}: A concept-annotated variant of CIFAR-10, comprising 26000 images, 18 concepts, and 10 object classes.
\end{itemize}

In all experiments, we adopted the sequential training scheme. First, a CNN backbone (Inception-V3 for CUB, ResNet50 for Derm7pt and RIVAL10) was trained to predict concept scores from input images. Then, using the predicted concept vectors as inputs, we trained our CLPC classifier and compared it against a baseline logistic regression (LR) model. We evaluated the models along the following three dimensions:
\begin{enumerate}
    \item \textbf{Label prediction performance}: assessing the overall label prediction performance based on concept predictions.
    \item \textbf{Robustness to concept noise}: evaluating the model’s stability when predicted concept scores are corrupted by random perturbations.
    \item \textbf{Concept intervention efficiency}: measuring how effectively the model can be guided toward a desired class via minimal edits to the concept vector.
\end{enumerate}
Detailed settings, results, and analyses for each of these aspects are presented in the following subsections.

\subsection{Label prediction }

This experiment consists of two parts: single-label prediction and conformal prediction. For single-label prediction, we evaluate model performance using the standard Top-1 accuracy, which measures the proportion of samples for which the predicted class exactly matches the ground truth. For conformal prediction, which may output a set of candidate classes rather than a single label, we use three complementary metrics:
\begin{itemize}
    \item \textbf{Set Accuracy}: The proportion of prediction sets that contain the true label.
    \item \textbf{Average Set Size}: The average number of class labels included in the output set, reflecting prediction specificity.
    \item \textbf{Reject Ratio}: The percentage of predictions for which the model returns an empty set, indicating uncertainty too high to make a reliable prediction.
\end{itemize}

For our model, the nonconformity score is derived from the distance between the predicted concept vector and each class-level prototype:
$s_j(\boldsymbol{x}) = d(\hat{\boldsymbol{c}},\ \boldsymbol{p}_j),\ j=1,\dots , L$. For LR, we adopt $s_j(\boldsymbol{x})= 1 - \hat{p}(y_j \,|\, \hat{\boldsymbol{c}})$ as the nonconformity score where $\hat{p}(y_j \,|\, \hat{\boldsymbol{c}})$ is the evaluation of posterior probability for class $y_j$. We set the significance level $\alpha$ to 0.05 for all datasets.

\begin{table}[t]
    \centering
    \caption{Label prediction performance of compared models.}
    \renewcommand{\arraystretch}{1.5}
    \begin{tabular}{p{1.2cm}>{\centering\arraybackslash}p{2cm}*{4}{>{\centering\arraybackslash}p{1cm}}*{2}{>{\centering\arraybackslash}p{0.8cm}}*{2}{>{\centering\arraybackslash}p{1.3cm}}}
    \toprule[1.5pt]
        \multirow{2}{*}{Dataset} & \multirow{2}{*}{Concept Acc}&  \multicolumn{2}{c}{Top-1 Acc (\%)} & \multicolumn{2}{c}{Set Acc (\%)} & \multicolumn{2}{c}{Set Size} & \multicolumn{2}{c}{Reject Ratio (\%)} \\
    \cline{3-10}
     & & LR & CLPC & LR & CLPC & LR & CLPC & LR & CLPC \\
    \midrule
    CUB & 94.86 & 76.46 & 76.01 & 92.12 & 94.97 & 1 & 1 & 29.5 & 53.30 \\
    Derm7pt & 88.38 & 66.33 & 64.81 & 87.34 & 94.43 & 2.15 & 3.38 & 0 & 0 \\
    RIVAL10 & 99.71 & 99.17 & 98.96 & 99.96 & 99.92 & 1 & 1 & 5.07 & 5.37 \\
    \bottomrule[1.5pt]
    \end{tabular}
    \label{tab:label-accuracy}
\end{table}

As shown in Table~\ref{tab:label-accuracy}, our CLPC model achieves comparable Top-1 accuracy to LR across all datasets, indicating that prototype-based reasoning does not sacrifice classification performance. The performance gap is minimal on CUB and RIVAL10, while CLPC slightly underperforms on Derm7pt.

In the conformal prediction setting, CLPC achieves higher set accuracy on two datasets out of three, demonstrating its ability to provide valid uncertainty-aware predictions. This improvement comes with slightly larger set sizes, reflecting a more conservative prediction approach. CLPC also exhibits a higher reject ratio, especially on CUB, whose test accuracy is significantly lower than training accuracy, showing it is more likely to abstain when concept evidence is insufficient or over-observed, consistent with its reliance on prototype matching.

Based on these analyses, CLPC provides reliable predictive coverage and enhanced interpretability (decomposition of uncertainty) while maintaining accuracy, making it a strong alternative to conventional classifiers in concept-based prediction scenarios.

\subsection{Robustness to noise in concept prediction}

\begin{figure}[t]
    \centering

    \begin{subfigure}[b]{0.32\textwidth}
        \includegraphics[width=\textwidth]{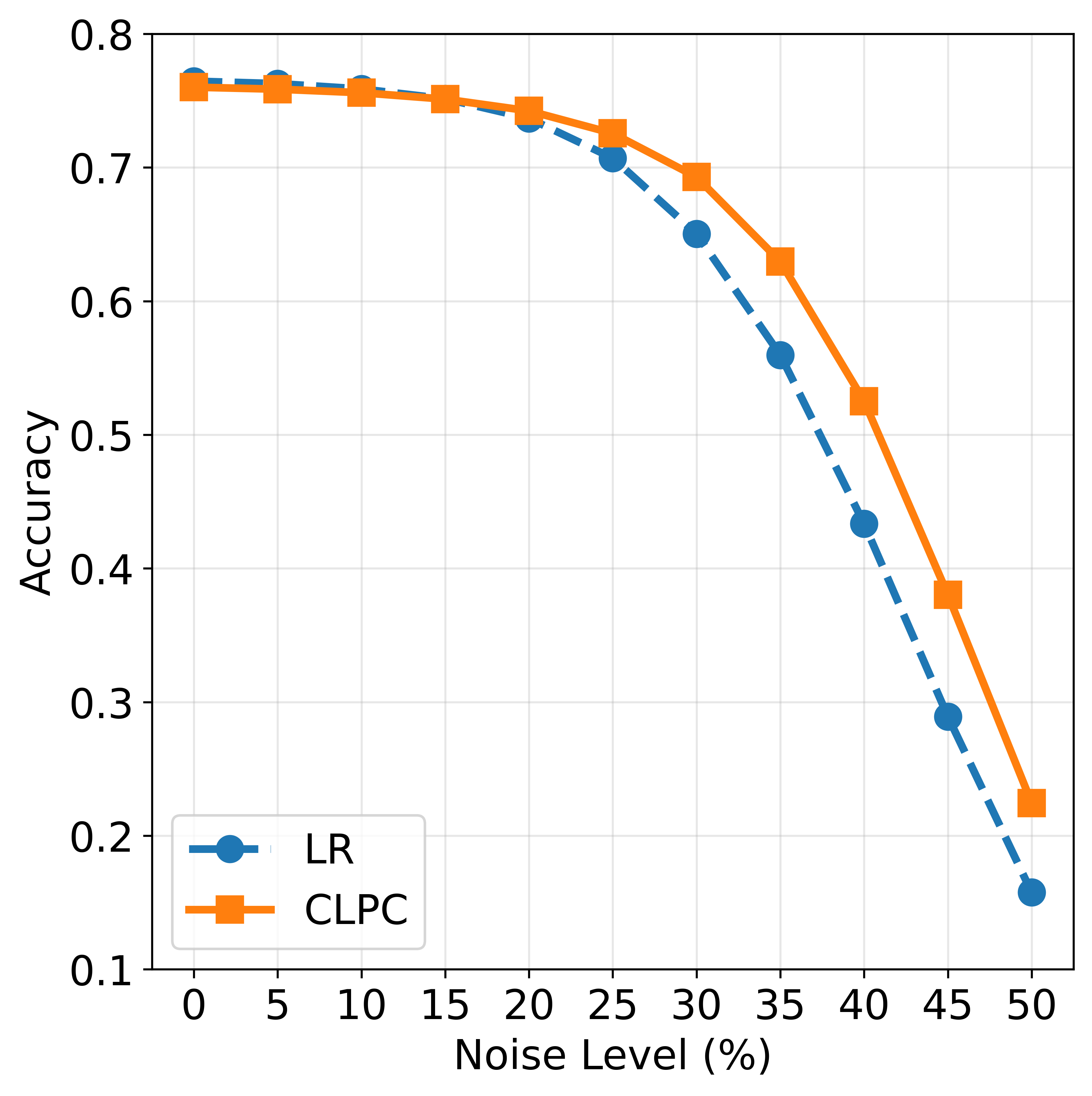}
        \caption{CUB}
        \label{fig:CUB-acc-wrt-noise}
    \end{subfigure}
    \hfill
    \begin{subfigure}[b]{0.32\textwidth}
        \includegraphics[width=\textwidth]{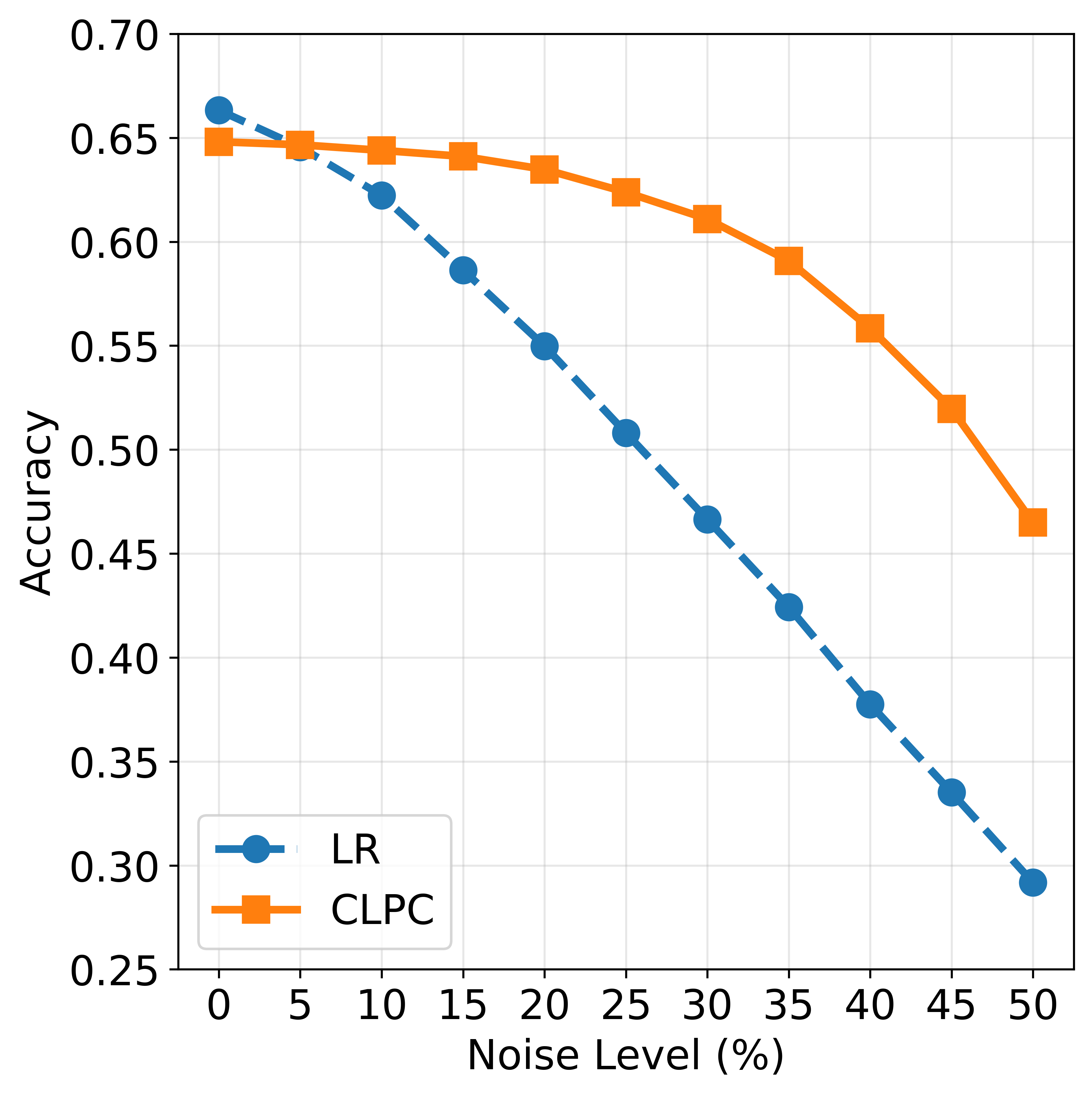}
        \caption{Derm7pt}
        \label{fig:Derm7pt-acc-wrt-noise}
    \end{subfigure}
    \hfill
    \begin{subfigure}[b]{0.32\textwidth}
        \includegraphics[width=\textwidth]{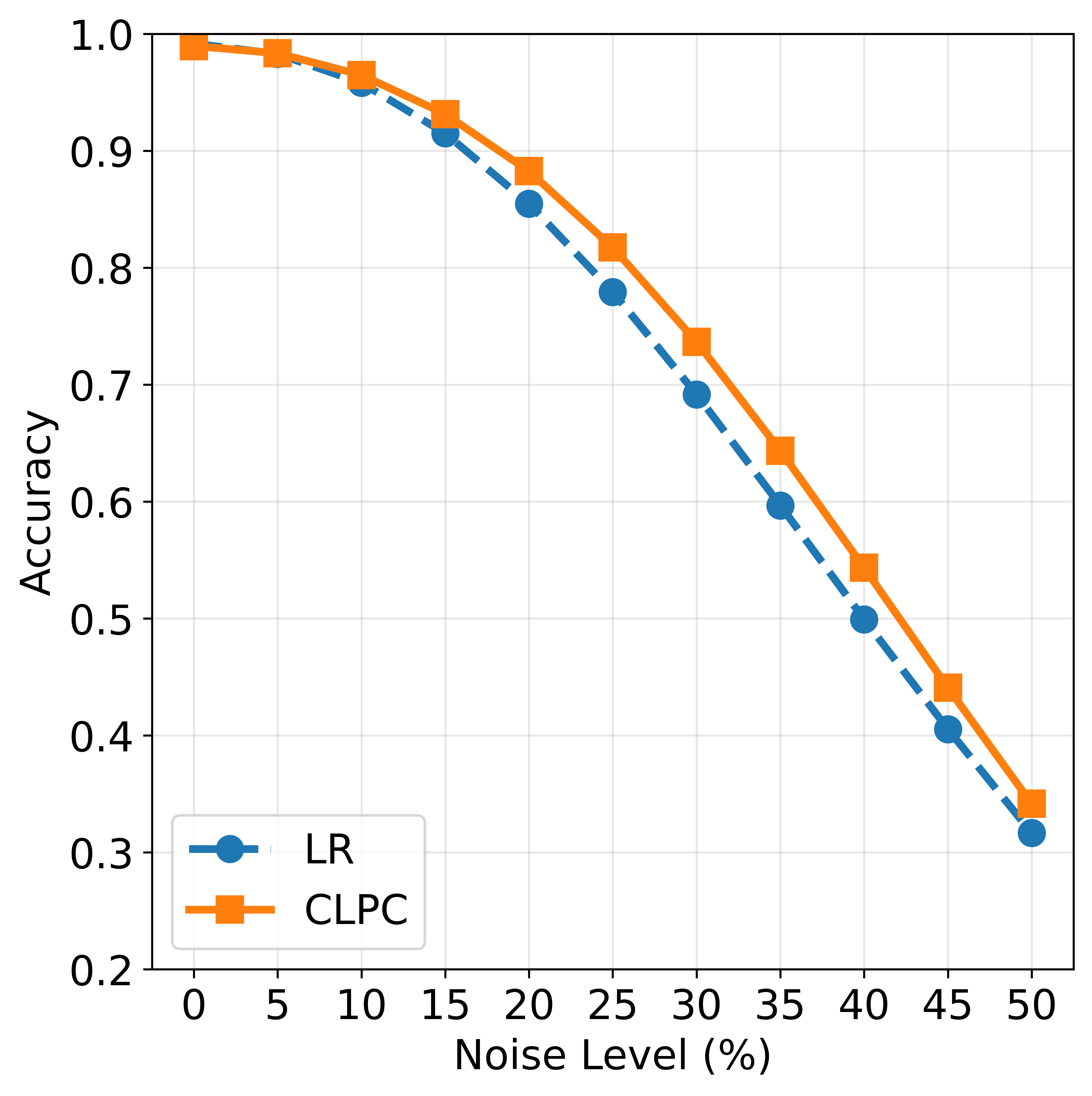}
        \caption{RIVAL10}
        \label{fig:RIVAL10-acc-wrt-noise}
    \end{subfigure}

    \caption{Top-1 accuracy with respect to levels of concept noise in test set.}
    \label{fig:acc-wrt-noise}
\end{figure}

In this experiment, we investigated the model's robustness to noise in the concept predictions from the first stage. Such noise may arise when the input image is unclear or the training data is insufficient. To simulate this, we fixed the trained model and varied the noise level from 0\% to 50\% on the test set. For a given noise level $t\%$, and for each test instance, we randomly selected $t\%$ of the concepts. For each selected concept, if its predicted score was below 0.5, we increased it to a random value between 0.5 and 1; otherwise, we decreased it to a random value between 0 and 0.5. For each dataset, we repeated this process 100 times and reported the average Top-1 accuracy in Fig.~\ref{fig:acc-wrt-noise} to illustrate the model's robustness to concept prediction noise.

As shown in Fig.~\ref{fig:acc-wrt-noise}, both CLPC and LR show decreasing Top-1 accuracy as the noise level increases. However, CLPC consistently outperforms LR across all datasets, particularly under moderate to high noise conditions. This improved robustness may stem from the nature of the prediction mechanism in CLPC, which compares the entire concept vector to each class-level prototype and selects the class with the smallest distance. Although both CLPC and LR consider all concepts, LR relies on a weighted sum of concept scores, which can amplify the effect of noise in highly weighted concepts. In contrast, CLPC evaluates the match between the whole predicted concept vector and the expected concept configuration of each class. This distance-based aggregation may provide a more balanced and stable criterion under noisy conditions, as errors in individual concepts are less likely to dominate the final decision.

\subsection{Concept intervention efficiency}

\begin{figure}[t]
    \centering

    \begin{subfigure}[b]{0.32\textwidth}
        \includegraphics[width=\textwidth]{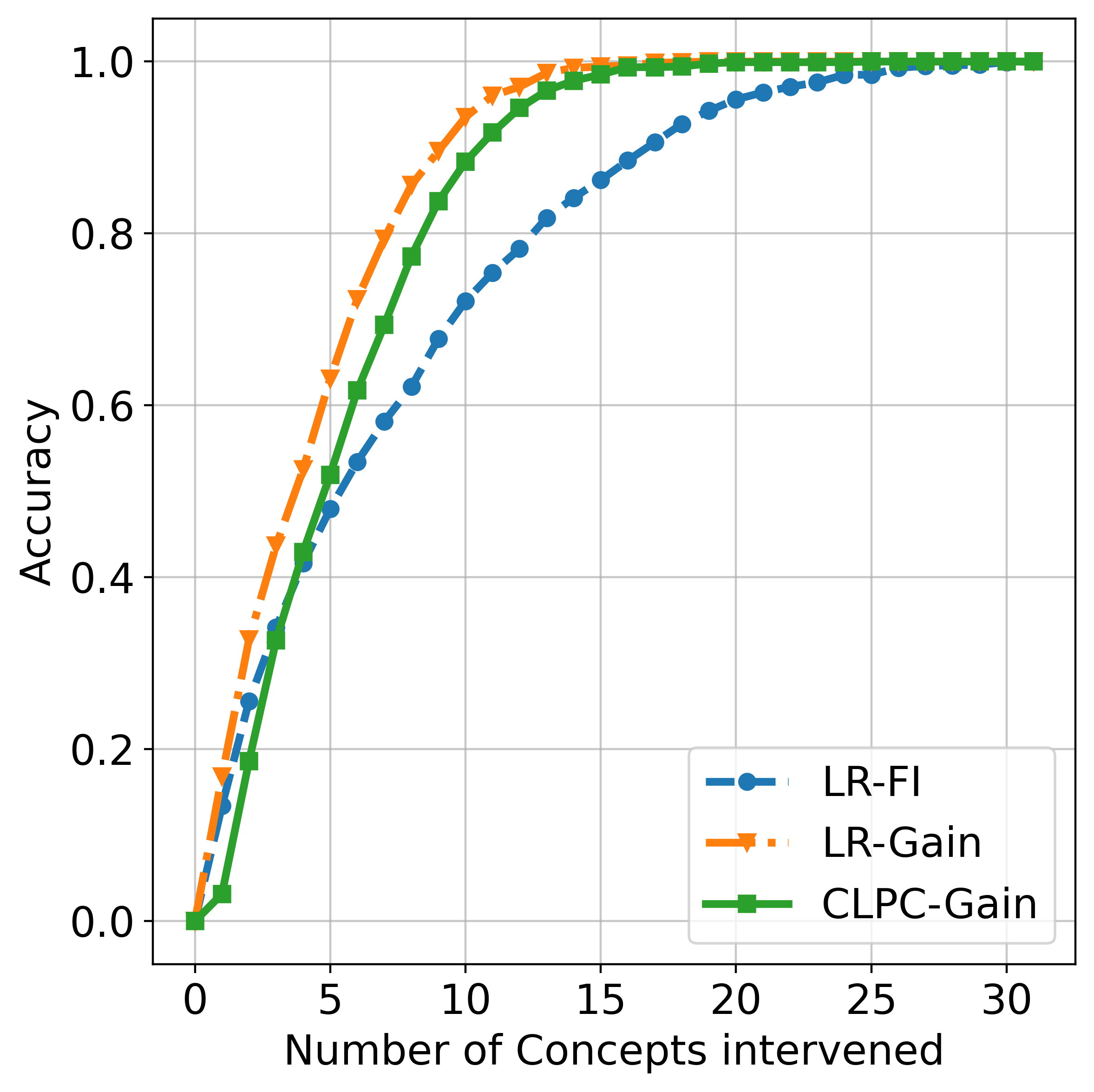}
        \caption{CUB}
        \label{fig:CUB-intervention-efficiency}
    \end{subfigure}
    \hfill
    \begin{subfigure}[b]{0.32\textwidth}
        \includegraphics[width=\textwidth]{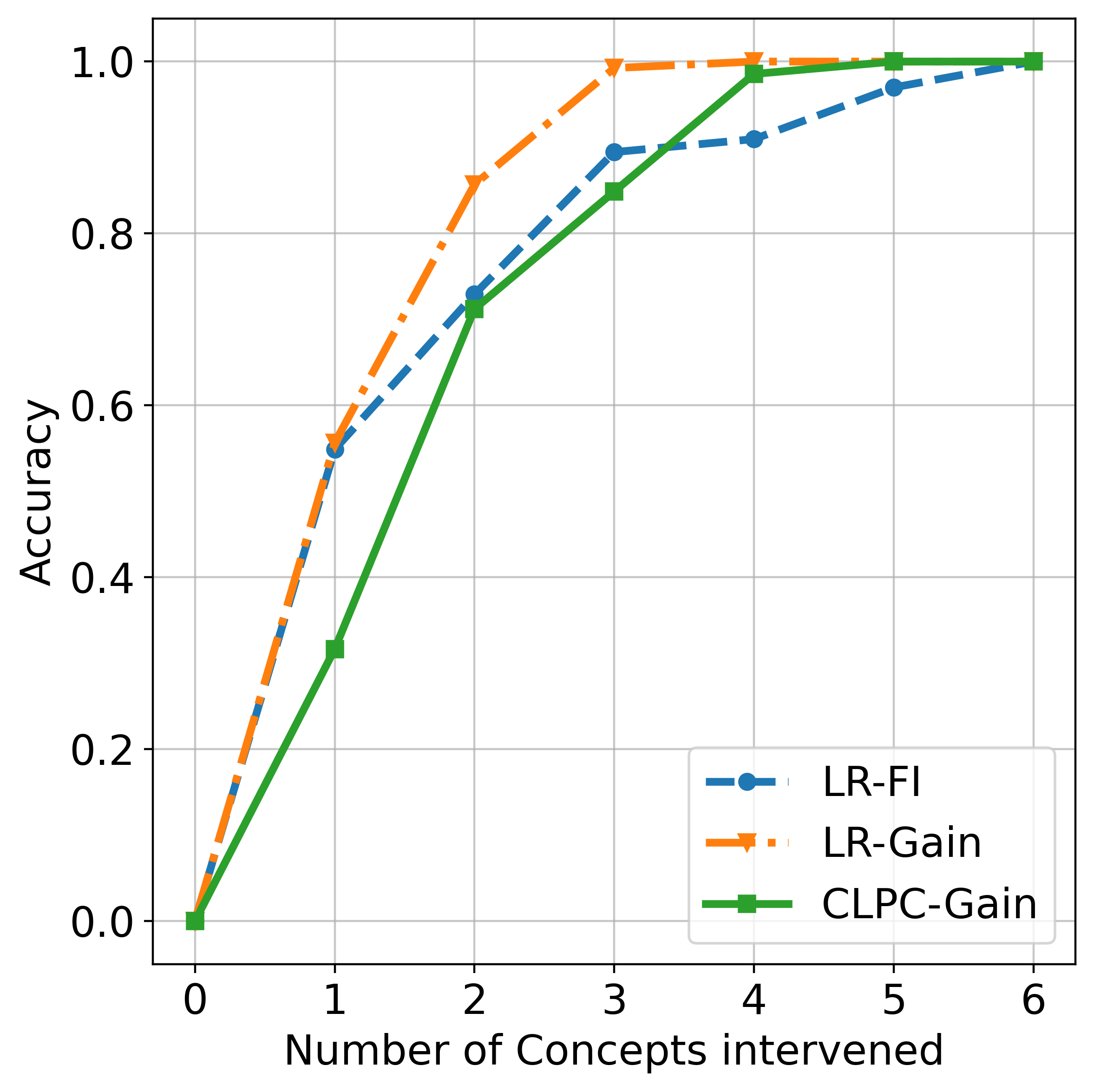}
        \caption{Derm7pt}
        \label{fig:Derm7pt-intervention-efficiency}
    \end{subfigure}
    \hfill
    \begin{subfigure}[b]{0.32\textwidth}
        \includegraphics[width=\textwidth]{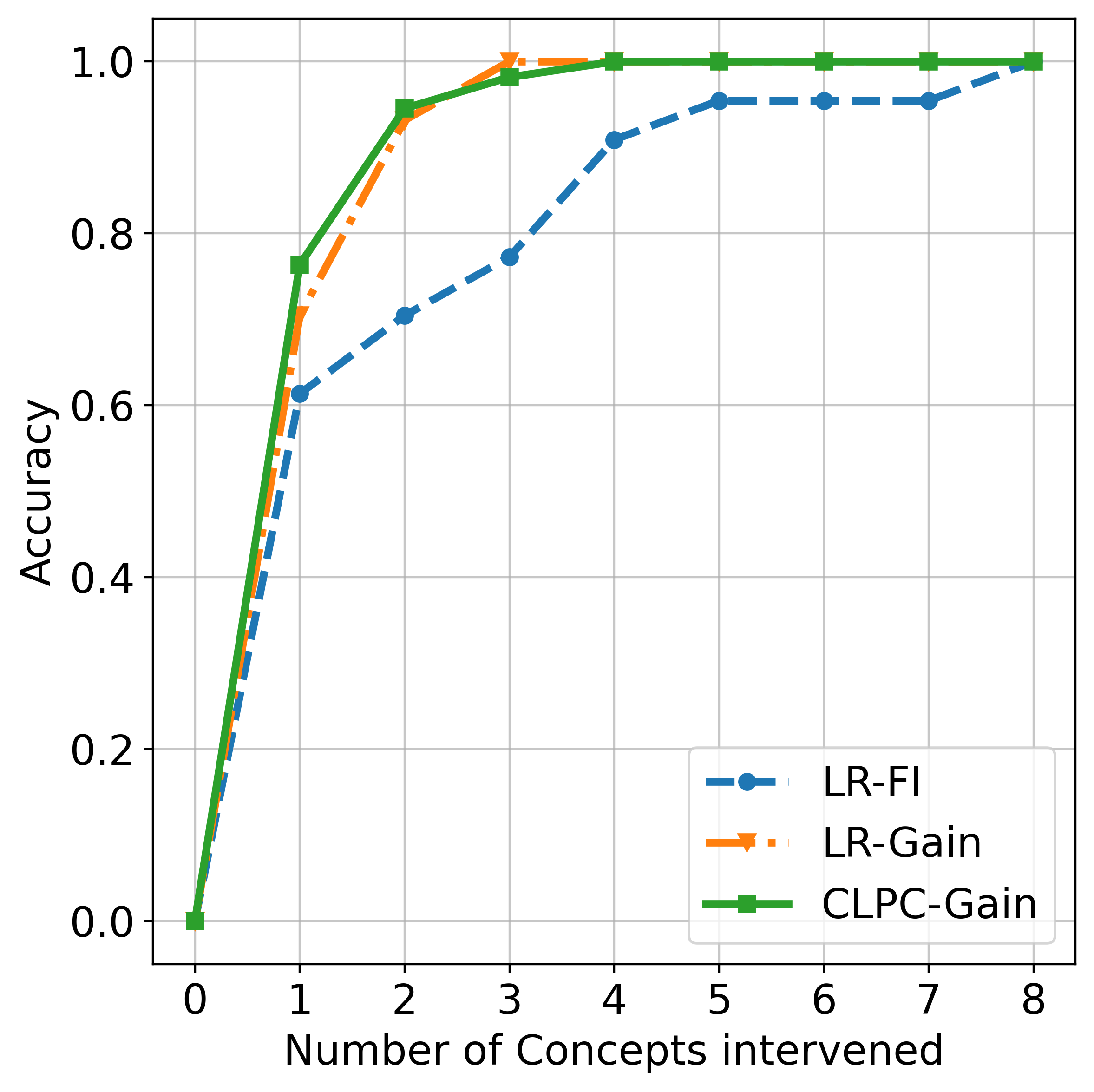}
        \caption{RIVAL10}
        \label{fig:RIVAL10-intervention-efficiency}
    \end{subfigure}

    \caption{Number of concept interventions required to correct misclassified test instances.}
    \label{fig:intervention-efficiency}
\end{figure}

In this section, we conduct a targeted correction experiment on misclassified samples to evaluate the effectiveness of different concept intervention strategies. For each misclassified instance $\boldsymbol{x}$, the goal is to flip the model’s prediction to the correct class $y$ by iteratively correcting predicted concept scores in $\boldsymbol{\hat{c}}$. At each step, one concept $\hat{c}_k$ is corrected based on a predefined priority ranking, and the model's label prediction is updated accordingly. This process continues until the prediction matches the ground-truth label, and we record the number of concepts intervened. For priority ranking, we compare two strategies in the logistic regression setting: LR-FI, which uses feature importance based on absolute weights, and LR-Gain, which applies the potential gain criterion defined in Eq.~\eqref{eq:lt-gain}. In our CLPC model, where feature importance is not explicitly defined, we evaluate only the potential gain-based strategy, denoted as CLPC-Gain, which prioritizes concepts by their alignment with the target prototype (see Eq.~\eqref{eq:clpc-gain}).

The results in Fig.~\ref{fig:intervention-efficiency} demonstrate that potential gain-based strategies (LR-Gain and CLPC-Gain) consistently outperform the feature importance baseline (LR-FI), requiring fewer concept corrections to flip misclassified predictions. This highlights the advantage of incorporating both the model's learned parameters and the current concept prediction state, rather than relying solely on static feature importance scores.

As for the comparison between LR-Gain and CLPC-Gain, we can find that CLPC-Gain is slightly less efficient. This is consistent with our finding in the experiment of robustness to noise, which can be explained as that CLPC is less sensitive to the modification in concept prediction scores than LR. However, their concept-intervention efficiency is quite comparable.

\section{Conclusion}\label{sec:5}

In this work, we proposed a novel class-level prototype classifier (CLPC) for Concept Bottleneck Models (CBMs), addressing the dual challenges of interpretability and uncertainty in concept-to-label mappings. By representing each class through a binary prototype in the concept space and computing prediction scores via distances to these prototypes, our approach provides an intuitive and semantically meaningful classification mechanism. Furthermore, we demonstrated how these distances can be decomposed into per-concept contributions, enabling fine-grained local explanations, prototype-based global interpretability, and effective concept interventions.

Experiments on CUB, Derm7pt, and RIVAL10 show that CLPC achieves comparable label accuracy to logistic regression, while offering better uncertainty calibration via higher set accuracy and conservative abstention. CLPC is also more robust to noisy concept predictions, benefiting from its distance-based reasoning. Although slightly less efficient in concept intervention than LR, CLPC still supports effective guided corrections. These results highlight CLPC as a reliable and interpretable alternative for logistic regression classifiers in CBMs.

For future work, we will compare our model with other uncertainty-aware CBMs, such as probabilistic CBM \cite{kim2023probabilistic} and Stochastic CBM \cite{vandenhirtz2024stochastic}. We will also explore the application of other uncertainty measurement frameworks in CBM, not only for concept-to-label mapping but also for image-to-concept encoding.

\begin{credits}
\subsubsection{\ackname} This work is funded by the Funds for New Lecturers of Jean Monnet University (Grant Number G752UJMIS3).

\subsubsection{\discintname}
The authors have no financial or personal relationships that could inappropriately influence or bias the content of the paper.

\subsubsection{Data Availability.}
Datasets, code, and experimental results for this paper are available on \href{https://github.com/patbarry29/clpc-cbm-model}{https://github.com/patbarry29/clpc-cbm-model}.
\end{credits}

%
%
%
\printbibliography
\end{document}